\documentclass[conference]{IEEEtran}
\IEEEoverridecommandlockouts
\usepackage{cite}
\usepackage{amsmath,amssymb,amsfonts}
\usepackage{algorithmic}
\usepackage{graphicx}
\usepackage{textcomp}
\usepackage{xcolor}
\usepackage{multirow}
\usepackage{booktabs}
\usepackage{soul}
\usepackage{color,xcolor}
\usepackage{marginnote}
\usepackage{todonotes}
\def\BibTeX{{\rm B\kern-.05em{\sc i\kern-.025em b}\kern-.08em
    T\kern-.1667em\lower.7ex\hbox{E}\kern-.125emX}}
\begin{document}

\title{EIAD: Explainable Industrial Anomaly Detection Via Multi-Modal Large Language Models}

\author{
    \IEEEauthorblockN{
        Zongyun Zhang\textsuperscript{1},
        Jiacheng Ruan\textsuperscript{1}, 
        Xian Gao\textsuperscript{1}, 
        Ting Liu\textsuperscript{1}, 
        Yuzhuo Fu\textsuperscript{1,\dag}\thanks{\textsuperscript{\dag}corresponding author}}
\IEEEauthorblockA{\textit{\textsuperscript{1}Shanghai Jiao Tong University}\\
\{zy.zhang2024, jackchenruan, gaoxian, louisa\_liu, yzfu\}@sjtu.edu.cn}
}

\maketitle

\begin{abstract}
Industrial Anomaly Detection (IAD) is critical to ensure product quality during manufacturing. Although existing zero-shot defect segmentation and detection methods have shown effectiveness, they cannot provide detailed descriptions of the defects. Furthermore, the application of large multi-modal models in IAD remains in its infancy, facing challenges in balancing question-answering (QA) performance and mask-based grounding capabilities, often owing to overfitting during the fine-tuning process. To address these challenges, we propose a novel approach that introduces a dedicated multi-modal defect localization module to decouple the dialog functionality from the core feature extraction. This decoupling is achieved through independent optimization objectives and tailored learning strategies. Additionally, we contribute to the first multi-modal industrial anomaly detection training dataset, named Defect Detection Question Answering (DDQA), encompassing a wide range of defect types and industrial scenarios. Unlike conventional datasets that rely on GPT-generated data, DDQA ensures authenticity and reliability and offers a robust foundation for model training. Experimental results demonstrate that our proposed method, Explainable Industrial Anomaly Detection Assistant (EIAD), achieves outstanding performance in defect detection and localization tasks. It not only significantly enhances accuracy but also improves interpretability. These advancements highlight the potential of EIAD for practical applications in industrial settings. Code and datasets will be available at https://github.com/Solunny/EIAD.
\end{abstract}


\section{Introduction}
\label{sec:intro}

Industrial anomaly detection plays a pivotal role in ensuring the reliability and safety of modern manufacturing systems. As industrial processes become increasingly complex and the diversity of potential defects grows, existing methods\cite{unsuper1,unsuper3} that primarily rely on predicting pixel-level anomaly scores are limited in their ability to define and describe anomalous regions precisely. Interpretative details related to domain knowledge such as anomaly categories, visual semantics, and potential causes and consequences are often overlooked. Recently, large multi-modal models have shown promising capabilities across various domains by integrating visual and language understanding to solve complex tasks. However, their application in industrial anomaly detection remains in its infancy, and several critical challenges remain to be addressed.

One of the most significant obstacles is achieving an effective balance between question-answering (QA) performance and grounding capabilities. Current multi-modal defect detection frameworks often suffer from overfitting during grounding-supervised fine-tuning, which limits their generalization across diverse industrial scenarios. This limitation becomes particularly pronounced when addressing the complexity and variability of real-world defect patterns.

To overcome these challenges, we constructed the first large-scale, rule-based training dataset specifically designed for industrial anomaly detection. Unlike conventional datasets that mostly rely on synthetic data generated by large models, our dataset is created based on existing annotations and domain-specific rules to ensure authenticity and reliability. This approach reduces the data noise caused by hallucinations, significantly lowers dataset construction costs, and captures the diversity of industrial environments by encompassing a wide range of tasks and defect types.

In addition, we propose a novel industrial anomaly detection method, \textbf{E}xplainable \textbf{I}ndustrial \textbf{A}nomaly \textbf{D}etection assistance (\textbf{EIAD}). This system introduces a dedicated multi-modal defect localization module to decouple the dialog functionality from the grounding outputs. By designing independent optimization objectives, our approach improves both detection accuracy and interpretability. This framework culminates in an integrated pipeline for defect analysis, detection, and localization.

Extensive experiments demonstrate that EIAD achieves outstanding performance in defect detection and localization tasks, highlighting its potential for practical industrial applications.
Our contributions are summarized as follows:
\begin{itemize}
  \item We developed the first dataset specifically tailored for industrial anomaly detection, called \textbf{DDQA-set} (\textbf{D}efect \textbf{D}etection \textbf{Q}uestion-\textbf{A}nswering dataset). This dataset spans four diverse tasks, encompassing various defect types and scenarios. Unlike traditional datasets, we avoided synthetic data generation with pretrained models such as GPT, ensuring data authenticity and reliability.
  \item Our approach decouples dialog and grounding capabilities by introducing a dedicated multi-modal defect localization module, supported by independent optimization objectives and a staged learning strategy to mitigate overfitting and enhance generalization.
  \item Extensive experiments demonstrate that our EIAD achieves superior performance in both defect localization and language interaction, compared to existing methods.
\end{itemize}

\begin{figure*}[!htbp]
\centering
\includegraphics[width=2.0\columnwidth]{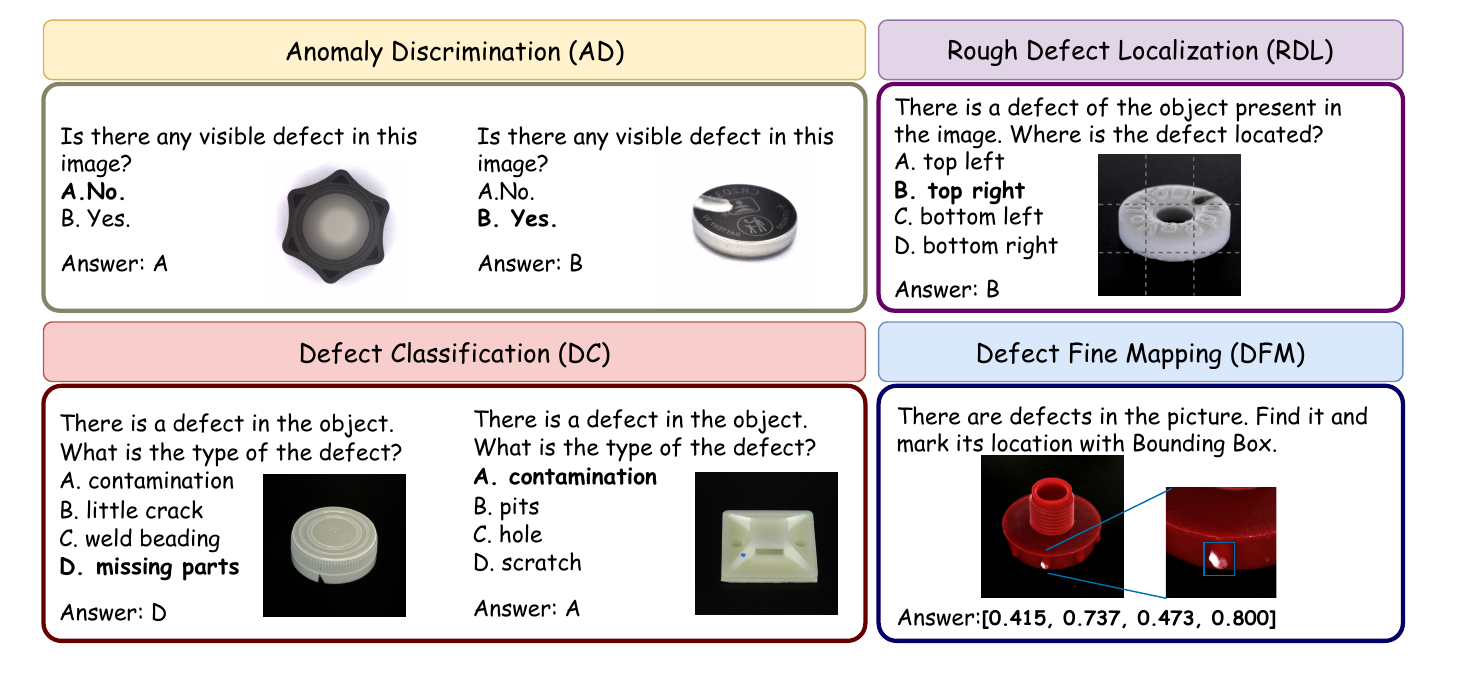} 
\caption{Examples of four tasks from the DDQA-set.}
\label{tasks}
\end{figure*}

\section{Related Work}
\subsection{Industrial Anomaly Detection (IAD)}
Industrial anomaly detection (IAD) has long been a critical area of research due to its essential role in quality control and the reliable operation of complex industrial systems. Traditional methods typically require extensive datasets of normal images to train models for each category, making the process time-intensive and costly.
To address these limitations, zero-shot IAD approaches leverage multi-modal foundation models to adapt to diverse defect categories without requiring reference images. For instance, AnomalyCLIP\cite{anomalyclip} employs object-agnostic text prompts and a combined global-local context optimization strategy to effectively capture normal and abnormal semantics.
Despite their advancements, these methods predominantly rely on predicted pixel-level anomaly scores, limiting their ability to precisely delineate and describe anomaly regions. Furthermore, they often overlook explanatory details grounded in domain knowledge, such as anomaly categories, visual semantics, and insights into potential causes and consequences.
To address these challenges, we incorporating large vision-language models (LVLMs) for Industrial Anomaly Segmentation (IAS) to gain a more comprehensive understanding of anomalous regions. 


\subsection{Large Visual language models (LVLMs)}
Large Vision-Language Models (LVLMs) integrate visual modalities into large language models and have demonstrated exceptional perception and generalization capabilities across a wide range of vision tasks including image captioning, visual question answering and cross-modal retrieval\cite{llava,gpt4v-ceping}. Recent studies have increasingly focused on enhancing the fine-grained grounding capabilities of LVLMs. For instance, methods such as LISA\cite{lisa}, GLaMM\cite{glamm}, and SAM4MLLM\cite{sam4mllm} incorporate specific vocabularies into models and leverage decoding modules to generate mask grounding outputs. These approaches typically employ grounding supervision during fine-tuning to refine the localization performance of LVLMs further.

\subsection{LVLM based IAD methods}

AnomalyGPT\cite{anomalygpt} is the first approach to incorporate large vision-language models (LVLMs) into industrial anomaly detection by leveraging an LLM-based image decoder to generate anomaly maps. However, it fails to fully exploit the visual understanding capabilities of large multi-modal models, resulting in suboptimal text generation quality. Myriad\cite{myriad} improves upon AnomalyGPT by integrating domain knowledge as embeddings based on textual and visual features, utilizing these embeddings to extract aligned vision-language representations. Nevertheless, it still lacks explicit domain knowledge data in the form of direct language inputs.

In contrast, our work addresses these limitations by pursuing the following objectives: (1) Constructing question-answering instruction data tailored to the industrial anomaly detection (IAD) domain and using this data, together with existing fine-grained grounding labels, as supervisory signals;
(2) Performing staged fine-tuning of the language model and defect localization module separately, thereby avoiding interference caused by multi-task training.

\section{Methodology}
\subsection{Construction of the Proposed DDQA-set}
\paragraph{Motivation}
Existing defect detection datasets primarily focus on visual mask annotations and defect class labels, lacking the semantic information necessary for training large vision-language models (LVLMs). To address this issue, we transform the traditional grounding information and class labels into instruction formats specifically designed for LVLM training. Based on empirically derived rules, we constructed the first multi-modal training dataset tailored for defect detection tasks, providing a robust foundation for future research in this domain. 

This method offers several significant advantages. Firstly, unlike approaches that rely on automated data generation using models like GPT-4, our method avoids noise introduced by issues such as model hallucinations and inherent biases, ensuring higher data quality and semantic accuracy. Moreover, the proposed approach substantially reduces the cost of data generation compared to traditional methods, achieving greater efficiency in producing multi-modal data while maintaining reliability and precision.

\paragraph{Data Collection}
We adopt 4 industrial anomaly detection datasets, with a total of 127997 images for DDQA-set. These datasets cover a variety of object categories and complex scenes, which are listed below.

\textbf{VISION}\cite{vision} gathers 14 datasets from real-world industrial detection scenarios spanning a variety of manufacturing processes, materials and industrial fields. 

\textbf{PR-REAL}\cite{pd-real} comprises RGB and depth images of Play-Doh models with six anomaly types, designed for 3D anomaly detection. 

\textbf{Real-IAD}\cite{real-iad} is a large-scale, multi-view industrial anomaly detection dataset with 15,000 samples across 30 categories, including materials like plastic, wood, ceramic, and composites.

\textbf{BSData} is an image dataset for ball screw drive containing 1104 images of RGB ball screw drive and 394 image masks related to surface damage.

\paragraph{Task Construction}
As shown in figure \ref{tasks}, we designed four anomaly-related tasks to enhance the capabilities of the model in industrial quality inspection. All data is constructed based on existing information in the dataset, including images, defect masks, object and defect categories, following specific rules. No generative models, such as GPT-4, are utilized in the data construction process, effectively reducing data noise and increasing accuracy and credibility. The statistical information of the datasets is shown in table \ref{train_dataset}.

\begin{table}[htb]
    \centering
    \caption{Statistical information of DDQA-set.}\label{train_dataset}
    \begin{tabular}{@{}cccccc@{}}
    \toprule
    \multirow{2}{*}{Task} &\multirow{2}{*}{Questions} &\multicolumn{4}{c}{Dataset}  \\
    && Real-IAD&PD-REAL&VISION&BSData\\ \midrule

    AD & 126103& 1104 &830 &124169& 0\\
    
    RDL   & 53124& 394 & 440 & 51329 & 961 \\
    
    DFM  & 47281& 331 & 511 & 45309 & 1130 \\
    
    DC    & 54147& 394 & 530 & 51329 & 1894 \\

    \cmidrule{1-6}
    Total & 280655 & 2223 & 2311 & 272136 & 3985 \\
     \bottomrule
    \end{tabular}
\end{table}

\begin{itemize}
\item \textbf{Anomaly Discrimination (AD)}: This is a binary classification task asking the model to determine whether a defect exists in a given sample. The objective is to improve the model's ability to detect the presence of anomalies.
\item \textbf{Rough Defect Localization (RDL)}: This task requires the model to identify the approximate location of defects. Images are divided into a 3x3 grid of nine equal regions, named as the top left corner, top right corner, etc. This task evaluates the performance of the model in coarse-grained localization.
\item \textbf{Defect Fine Mapping (DFM)}: For this task, the model must pinpoint the precise location of defects. Answers are required to be provided in the form of bounding boxes, testing the model's fine-grained localization ability.
\item \textbf{Defect Classification (DC)}: This task challenges the model to classify the type of defect, assessing its semantic understanding of various industrial anomalies.
\end{itemize}



 To verify the difficulty of the dataset, we conducted comprehensive experiments on the proposed DDQA dataset using the MiniCPM-V2.6\cite{minicpm}. Results are shown in table \ref{dataset_compare}. Compared to MMAD\cite{mmad}, the only existing multi-modal defect detection benchmark, the same model performs lower accuracy on our proposed DDQA-set, indicating that DDQA has significantly higher difficulty, especially on defect localization tasks. The experimental results indicate that the accuracy of the DDQA model is close to that of random selection. Specifically, our introduced Defect Fine Mapping (DFM) task, which is not available in the MMAD dataset, imposes stricter requirements for precise defect localization. 

\begin{table}[!htbp]
    \centering
    \caption{Comparison of dataset difficulty on DDQA and MMAD.}\label{dataset_compare}
    \begin{tabular}{@{}cccccc@{}}
    \toprule
    \multirow{2}{*}{Model} &\multirow{2}{*}{Dataset} &\multicolumn{4}{c}{Tasks}  \\
    && AD& DC & RDL&Average\\ \midrule
    Random Chance&-&50.0&25.0&25.0&33.3 \\
    \cmidrule{1-6}
    \multirow{2}{*}{MiniCPM-V2.6}&MMAD & 53.4& 48.3&53.9&51.9\\
    
    &DDQA (ours)   & 65.1 & 27.2&  17.9&36.7\\

     \bottomrule
    \end{tabular}
    
\end{table}

\begin{figure*}[!htbp]
\centering
\includegraphics[width=2.0\columnwidth]{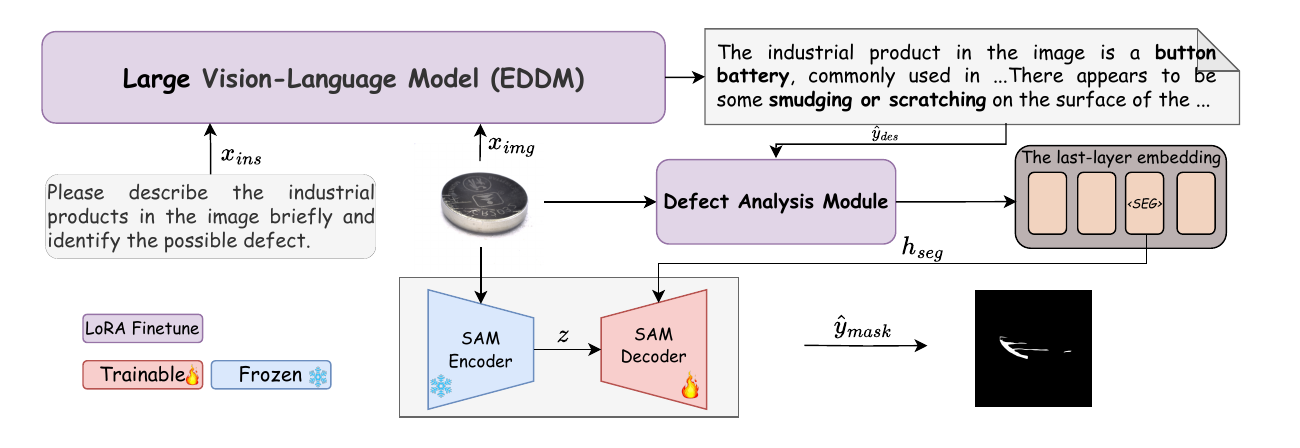}
\caption{The pipeline of EIAD. Given an image $x_{img}$ for detection, input it into the fine-tuned LVLM along with a specified text instruction $x_{ins}$ simultaneously to generate image description $\hat{y}_{des}$. Subsequently, $\hat{y}_{des}$ and $x_{ins}$ are input into the Defect Analysis Module, and the last-layer embedding for the [SEG] token $h_{seg}$ serves as a prompt for SAM, guiding it to generate the defect mask grounding $\hat{y}_{mask}$.}
\label{model}
\end{figure*}
\subsection{Overall Framework of EIAD}
Our approach fundamentally differs from prior methods in the design and handling of multitask joint training. Conventional approaches often employ multitask joint training to simultaneously optimize defect recognition and grounding outputs. However, the data used in such training are often incomplete and diverse, which can lead to interference among tasks during network optimization, particularly when overemphasizing one task at the expense of others. To address this issue, we introduce a decoupling strategy that independently manages task interactions, enabling the accumulation of visual priors and fine-grained perception to mutually reinforce each other and enhance overall performance.

Specifically, our industrial anomaly detection system (EIAD) consists of two primary modules: a large vision-language model and a dedicated segmentation module, named \textbf{E}xplainable \textbf{D}efect \textbf{D}etection \textbf{M}odule (\textbf{EDDM}) and \textbf{M}ulti-modal \textbf{D}efect \textbf{L}ocation \textbf{M}odule (\textbf{MDLM}) respectively. Our research objectives focus on two key aspects: 1) leveraging the multi-modal large language model for its textual understanding and world knowledge priors to analyze and infer potential defects in images; 2) utilizing semantic analysis and interpretation of images to assist the segmentation module in accurately localizing defect regions. This modular design significantly improves the model's performance and usability, providing an efficient and interpretable solution for industrial anomaly detection.

As illustrated in figure \ref{model}, given an image input $x_{img}$ and a textual query $x_{ins}$ (e.g., “Please briefly describe the industrial product in the image and identify potential defects”), they are first fed into EDDM module to generate semantic descriptions of the image and potential defect descriptions $\hat{y}_{des}$. Subsequently, the description $\hat{y}_{des}$ and the image $x_{img}$ are input into the MDLM module to accurately extract the defect mask $\hat{y}_{mask}$. To enhance cross-modal interactions and improve defect localization accuracy, we introduce a dedicated \textbf{D}efect \textbf{A}nalysis \textbf{M}odule (\textbf{DAM}) in MDLM. This module aligns visual and textual features while augmenting the visual backbone's capability to process complex and lengthy descriptions. This design strengthens the model's cross-modal understanding and significantly improves the precision and robustness of defect localization.

\paragraph{Multi-modal Defect Location Module}
The output of the previous module $\hat{y}_{des}$ provides a textual description of the image content and potential defects. The current objective is to use $\hat{y}_{des}$ and $x_{img}$ to obtain an accurate binary mask, offering a clearer and more precise representation of the defect regions.

Inspired by LISA\cite{lisa}, we propose a Defect Analysis Module (DAM) that aligns long textual features obtained from the previous module EDDM with visual modality features. The aligned features are then fed into the decoder of the segmentation model SAM\cite{sam} to generate a segmentation mask.

Specifically, in MDLM, the image is first input into the visual backbone to obtain a visual feature embedding $z$. Then the 'embedding-as-mask' paradigm is used to generate segmentation output. Specifically, the token of [SEG] is added to the vocabulary of the large language model as a special word to request for segmentation output. The last-layer embedding corresponding to [SEG] is extracted and projected to $h_{seg}$. Finally, $h_{seg}$ and the visual feature $z$ are sent to the SAM decoder\cite{sam} to generate the final segmentation mask $\hat{y}_{mask}$.

\begin{table*}[!htbp]
\centering
\caption{Performance comparison of both proprietary and open-source MLLMs in MMAD with the standard zero-shot setting. (* For the methods using 1-shot setting.)}
\begin{tabular}{c|c|c|cccc|cc|c}
\toprule
\multirow{2}{*}{Model}&\multirow{2}{*}{Scale} & Anomaly& \multicolumn{4}{c|}{Defect} & \multicolumn{2}{c|}{Object} & \multirow{2}{*}{Average}\\
    &                  &Discrimination&Classification&Localization&Description&Analysis&Classification & Analysis &  \\
\cmidrule{1-10}
Random Chance & - & 50.0 & 25.0 & 25.0 & 25.0 & 25.0 & 25.0 & 25.0 & 28.6 \\
\cmidrule{1-10}
Gemini-1.5-flash* & - & 58.6 & 54.7 & 49.1 & 66.5 & 82.2 & 91.5 & 79.7 & 68.9 \\
GPT-4o-mini* & - & 64.3 & 48.6 &38.8& 63.7 & 80.4 & 88.6 & 79.7 & 66.3 \\
GPT-4o* & - & \textbf{68.6} & \textbf{65.8} & 55.6 & \textbf{73.2} & \textbf{83.4} & \textbf{95.0} & \textbf{82.8} & \textbf{74.9} \\
\cmidrule{1-10}
AnomalyGPT* & 7B & \textbf{65.6} & 27.5 & 28.0 & 36.9 & 32.1 & 29.8 & 35.8 & 36.5 \\
Cambrian-1 & 8B & 55.6&32.5 & 35.4 & 43.5 & 49.1 & 78.2 & 67.2 & 51.6 \\
Qwen-VL-Chat* & 7B & 53.7 & 31.3 & 28.6 & 41.7 & 64.0 & 74.5 & 67.9 & 51.7 \\
LLaVA-1.5* & 13B & 50.0 & 38.8 & 46.2 & 58.2 & 73.1 & 73.6&71.0 & 58.7 \\
SPHINX & 7B & 53.1&39.9 & 52.3 & 51.0 & 71.2 & 85.1 & 73.1 & 60.0 \\
InternLM-XComposer2-VL* & 7B & 55.9&41.8 & 48.3 & 57.5 & 76.7&74.3 & 77.8 & 61.7 \\
InternVL2* & 8B &60.0& 43.9 & 47.9 & 57.6 & \underline{78.1} & 74.2 & 80.4 & 63.1 \\
LLaVA-NeXT* & 34B & 57.9&\underline{48.8} & 52.9 & \textbf{71.3} & \textbf{80.3} & 81.1 & 77.8 & 67.2 \\
\cmidrule{1-10}
MiniCPM-V2.6 & 8B & 53.4 & 48.3 & \underline{53.9} & 67.0 &\underline{78.1}& \textbf{91.8} & \textbf{82.8} & \underline{67.9} \\
\textbf{EIAD (ours)} & 8B & \underline{60.5} & \textbf{50.7} & \textbf{55.6} & \underline{67.8} & 76.9 & \textbf{91.8} & \underline{82.7} &\textbf{69.4}\\
\hline
\end{tabular}
\label{performance_language}
\end{table*}

\subsection{Mask Grounding Supervision}
Mask grounding is required as supervision to train the MDLM module. In total, we utilized mask annotations from seven industrial anomaly datasets for training. This includes four datasets from DDQA and the following three additional datasets:

\textbf{RSDD} is a Rail Surface Anomaly Detection Dataset containing 195 images of rail surface defects from high-speed and regular/heavy transport rails, marked by experienced inspectors.

\textbf{DAGM2007} is a synthetic dataset for texture surface defect detection, containing of 10 types. Each type is generated by different texture and defect models, simulating various situations that may occur in the real world.

\textbf{Eyecandies}\cite{eyecandies} includes synthetic 3D images of candies and cookies with challenges like complex textures and self-occlusion.

\subsection{Training Objectives}
The two sub-modules of EIAD are trained end-to-end separately. 
For the EDDM module, the constructed DDQA dataset is used for training to equip the model with domain knowledge for defect detection. The loss function in this part employs autoregressive cross-entropy loss for text generation, as defined in the following equations.
$$\mathcal{L}_{EDDM}= l_{ce}(\hat{y}_{txt},y_{txt})$$
For the MDLM module, we use mask grounding supervision to finetune the model. Following LISA\cite{lisa}, our training objective is the loss combining binary cross-entropy loss and segmentation DICE loss, with corresponding loss weights $\lambda_{bce}$ and $\lambda_{dice}$, as defined in the following equations.
$$\mathcal{L}_{MDLM}= \lambda_{bce} {l}_{bce}(\hat{y}_{mask} ,y_{mask})  + \lambda_{dice} {l}_{dice}(\hat{y}_{mask} ,y_{mask})$$

\section{Experiment}

\subsection{Evaluation metrics and datasets}
For anomaly segmentation performance evaluation, we employ three \textbf{pixel-level} evaluation metrics to assess EIAD's anomaly localization performance. That's \textbf{AUROC} (Area Under Receiver Operating Characteristic), \textbf{F1-max} (F1 score at the optimal threshold) and \textbf{AP} score (Area Under Average Precision), respectively.
EIAD's zero-shot industrial anomaly segmentation performance are evaluated on two widely used public benchmarks: MVTec AD\cite{mvtec-ad} and ViSA\cite{visa}. To verify the generalization performance of the model, we also tested on the 3D dataset MVTec 3D-AD\cite{mvtec3dad}.

To evaluate interpretability, we use the latest full-spectrum MLLMs benchmark MMAD\cite{mmad} dataset. Considering the diversity and unpredictability of industrial scenarios, all our experiments are conducted under a \textbf{zero-shot} setting, ensuring that the data in the test set is never seen during training.

\subsection{Implementation Details}
On the DDQA-Set, we initially fine-tuned the EDDM module using LoRA (rank=8, alpha=32), such as MiniCPM-V2.6\cite{minicpm}. 
Afterward, we fine-tune the MDLM module (based on LLavA-v1.6\cite{llava}) and Segment Anything Model\cite{sam} with LoRA (rank=16, alpha=16), training for 200 steps on the same hardware configuration, with a learning rate of 2e-3 and a batch size of 8.

\subsection{Comparison with LVLMs}
To comprehensively evaluate the quality of defect detection, we conducted experiments using the MMAD dataset proposed in\cite{mmad} to validate the effectiveness of our method. As shown in table \ref{performance_language}, our approach demonstrates superior performance across all tasks in the MMAD dataset and is comparable to larger open-source models and most of the closed-source models.

Notably, our method achieves the best results in both defect localization and defect classification tasks. These findings underscore the critical role of the DDQA dataset, which effectively guides the model to acquire domain-specific knowledge relevant to defect detection, thereby significantly enhancing its overall performance.

\subsection{Comparison with Anomaly Detection Methods}
To assess the model’s capability to locate defect regions, we conduct comparisons on the MVTec AD\cite{mvtec-ad} and ViSA\cite{visa} datasets. We compare our model with prior zero-shot IAD methods, selecting WinCLIP\cite{winclip}, SAA+\cite{saa}, AnoCLIP\cite{anovl} and CLIP Surgery\cite{clipsurgery}. It can be observed that EIAD has better performance in locating the anomalous regions than the baseline models.

\begin{table}[!htbp]
    \centering
    \caption{Zero-shot anomaly detection results on MVTec AD\cite{mvtec-ad} and ViSA\cite{visa}. The best-performing method is in bold and the second best score is underlined.}\label{compare_mvtecad}
    \begin{tabular}{@{}l|cc|cc@{}}
    \toprule
    \multirow{2}{*}{Methods} &\multicolumn{2}{c|}{MVTec AD} &\multicolumn{2}{c}{ViSA}
     \\
     
     &AUROC & F1-max &AUROC & F1-max \\  \midrule   
    
    WinCLIP&
    85.1&31.7&
    79.6&14.8\\
    
    SAA+&
    72.2&\textbf{37.8}&
    74.0& \textbf{27.1}\\

    AnoCLIP&
    \underline{86.6}&30.1&
    83.7&13.5\\

    CLIP Surgery&
    83.5&29.8&
    \underline{85.0}&15.2\\
    

    \cmidrule{1-5}
    
    \textbf{EIAD (ours)}
    &\textbf{87.8} &\underline{34.7}&
    \textbf{91.9} &\underline{18.7}\\

    \bottomrule
    \end{tabular}
\end{table}

     
     
     
    
    
    




To validate the generalization ability of our EIAD, we also use a 3D anomaly detection dataset, MVTec 3D-AD\cite{mvtec3dad}, to test the segmentation results of EIAD on its RGB images. We compared the our EIAD model with the baseline method WinCLIP. The results are presented in table \ref{compare_3d}.

\begin{table}[!htbp]
    \centering
    \caption{Zero-shot anomaly detection results on MVTec 3D-AD\cite{mvtec3dad}.The best-performing method is in bold.}\label{compare_3d}
    \begin{tabular}{@{}ll|ccc@{}}
    \toprule
    \multirow{2}{*}{Methods} &\multirow{2}{*}{Base Model}&\multicolumn{3}{c}{MVTec 3D-AD}\\
     &&AUROC & F1-max  &AP-seg \\  \midrule   
    
    WinCLIP&ViT-B/16+&
    91.2&10.3&5.3\\
    
    \cmidrule{1-5}
    
    \textbf{EIAD (ours)} &LLaVA (7B)& \textbf{94.8} & \textbf{27.4} & \textbf{20.3}\\

    \bottomrule
    \end{tabular}
\end{table}

\subsection{Ablation Study}
To specifically validate the role of the EDDM module, we conducted an ablation study in which the EDDM was removed.
In this setup, only a fixed prompt formed like ``Please recognize the defect in the image and output the segmentation mask" was provided to the MLDM module without any additional defect information generated by EDDM. 

As shown in table \ref{Ablation studies}, the absence of the EDDM module led to a measurable decline in the quality of abnormal segmentation. This result demonstrates that the detailed defect descriptions generated by the EDDM module significantly enhance the grounding output.

\begin{table}[!htbp]
\centering
\caption{Ablation studies on the zero-shot anomaly segmentation task with and without EDDM module.}
\label{Ablation studies}
    \begin{tabular}{l|cc|cc}
        \toprule
       Dataset & \multicolumn{2}{c|}{MVTec AD} & \multicolumn{2}{c}{ViSA} \\
       Setting& AUROC & F1-max & AUROC & F1-max \\
        \midrule
        EIAD \textit{w/o EDDM} & 87.0 & 33.9 & 90.7 &  18.5\\
        EIAD & \textbf{87.8} & \textbf{34.7} & \textbf{91.9} & \textbf{18.7}\\
        \bottomrule
    \end{tabular}
\end{table}

\section{Conclusion}
In this work, we apply large vision-language models to industrial defect detection. The proposed framework, EIAD, demonstrates exceptional performance in defect detection, offering comprehensive explanations and precise localization. Moreover, it exhibits strong generalization capabilities across a wide range of operational scenarios. However, there is still room for improvement in this work. For example, enriching the tasks and content of multi-modal datasets, as well as simplifying redundant parts of model structures.
\bibliographystyle{IEEEbib}
\bibliography{icme2025references}

\begin{thebibliography}{10}

\bibitem{unsuper1}
Paul Bergmann, Michael Fauser, David Sattlegger, and Carsten Steger,
\newblock ``Uninformed students: Student-teacher anomaly detection with discriminative latent embeddings,''
\newblock in {\em Proceedings of the IEEE/CVF conference on computer vision and pattern recognition}, 2020, pp. 4183--4192.

\bibitem{unsuper3}
Jonathan Pirnay and Keng Chai,
\newblock ``Inpainting transformer for anomaly detection,''
\newblock in {\em International Conference on Image Analysis and Processing}. Springer, 2022, pp. 394--406.

\bibitem{anomalyclip}
Qihang Zhou, Guansong Pang, Yu~Tian, Shibo He, and Jiming Chen,
\newblock ``Anomalyclip: Object-agnostic prompt learning for zero-shot anomaly detection,''
\newblock {\em arXiv preprint arXiv:2310.18961}, 2023.

\bibitem{llava}
Haotian Liu, Chunyuan Li, Qingyang Wu, and Yong~Jae Lee,
\newblock ``Visual instruction tuning,''
\newblock {\em Advances in neural information processing systems}, vol. 36, 2024.

\bibitem{gpt4v-ceping}
Yunkang Cao, Xiaohao Xu, Chen Sun, Xiaonan Huang, and Weiming Shen,
\newblock ``Towards generic anomaly detection and understanding: Large-scale visual-linguistic model (gpt-4v) takes the lead,''
\newblock {\em arXiv preprint arXiv:2311.02782}, 2023.

\bibitem{lisa}
Xin Lai, Zhuotao Tian, Yukang Chen, Yanwei Li, Yuhui Yuan, Shu Liu, and Jiaya Jia,
\newblock ``Lisa: Reasoning segmentation via large language model,''
\newblock {\em arXiv preprint arXiv:2308.00692}, 2023.

\bibitem{glamm}
Hanoona Rasheed, Muhammad Maaz, Sahal Shaji, Abdelrahman Shaker, Salman Khan, Hisham Cholakkal, Rao~M Anwer, Eric Xing, Ming-Hsuan Yang, and Fahad~S Khan,
\newblock ``Glamm: Pixel grounding large multimodal model,''
\newblock in {\em Proceedings of the IEEE/CVF Conference on Computer Vision and Pattern Recognition}, 2024, pp. 13009--13018.

\bibitem{sam4mllm}
Yi-Chia Chen, Wei-Hua Li, Cheng Sun, Yu-Chiang~Frank Wang, and Chu-Song Chen,
\newblock ``Sam4mllm: Enhance multi-modal large language model for referring expression segmentation,''
\newblock in {\em European Conference on Computer Vision}. Springer, 2025, pp. 323--340.

\bibitem{anomalygpt}
Zhaopeng Gu, Bingke Zhu, Guibo Zhu, Yingying Chen, Ming Tang, and Jinqiao Wang,
\newblock ``Anomalygpt: Detecting industrial anomalies using large vision-language models,''
\newblock in {\em Proceedings of the AAAI Conference on Artificial Intelligence}, 2024, vol.~38, pp. 1932--1940.

\bibitem{myriad}
Yuanze Li, Haolin Wang, Shihao Yuan, Ming Liu, Debin Zhao, Yiwen Guo, Chen Xu, Guangming Shi, and Wangmeng Zuo,
\newblock ``Myriad: Large multimodal model by applying vision experts for industrial anomaly detection,''
\newblock {\em arXiv preprint arXiv:2310.19070}, 2023.

\bibitem{vision}
Haoping Bai, Shancong Mou, Tatiana Likhomanenko, Ramazan~Gokberk Cinbis, Oncel Tuzel, Ping Huang, Jiulong Shan, Jianjun Shi, and Meng Cao,
\newblock ``Vision datasets: A benchmark for vision-based industrial inspection,''
\newblock {\em arXiv preprint arXiv:2306.07890}, 2023.

\bibitem{pd-real}
Jianjian Qin, Chunzhi Gu, Jun Yu, and Chao Zhang,
\newblock ``Image-pointcloud fusion based anomaly detection using pd-real dataset,''
\newblock {\em arXiv preprint arXiv:2311.04095}, 2023.

\bibitem{real-iad}
Chengjie Wang, Wenbing Zhu, Bin-Bin Gao, Zhenye Gan, Jianning Zhang, Zhihao Gu, Shuguang Qian, Mingang Chen, and Lizhuang Ma,
\newblock ``Real-iad: A real-world multi-view dataset for benchmarking versatile industrial anomaly detection,''
\newblock {\em arXiv preprint arXiv:2403.12580}, 2024.

\bibitem{minicpm}
Yuan Yao, Tianyu Yu, Ao~Zhang, Chongyi Wang, Junbo Cui, Hongji Zhu, Tianchi Cai, Haoyu Li, Weilin Zhao, Zhihui He, et~al.,
\newblock ``Minicpm-v: A gpt-4v level mllm on your phone,''
\newblock {\em arXiv preprint arXiv:2408.01800}, 2024.

\bibitem{mmad}
Xi~Jiang, Jian Li, Hanqiu Deng, Yong Liu, Bin-Bin Gao, Yifeng Zhou, Jialin Li, Chengjie Wang, and Feng Zheng,
\newblock ``Mmad: The first-ever comprehensive benchmark for multimodal large language models in industrial anomaly detection,''
\newblock {\em arXiv preprint arXiv:2410.09453}, 2024.

\bibitem{sam}
Alexander Kirillov, Eric Mintun, Nikhila Ravi, Hanzi Mao, Chloe Rolland, Laura Gustafson, Tete Xiao, Spencer Whitehead, Alexander~C Berg, Wan-Yen Lo, et~al.,
\newblock ``Segment anything,''
\newblock in {\em Proceedings of the IEEE/CVF International Conference on Computer Vision}, 2023, pp. 4015--4026.

\bibitem{eyecandies}
Luca Bonfiglioli, Marco Toschi, Davide Silvestri, Nicola Fioraio, and Daniele De~Gregorio,
\newblock ``The eyecandies dataset for unsupervised multimodal anomaly detection and localization,''
\newblock in {\em Proceedings of the Asian Conference on Computer Vision}, 2022, pp. 3586--3602.

\bibitem{mvtec-ad}
Paul Bergmann, Michael Fauser, David Sattlegger, and Carsten Steger,
\newblock ``Mvtec ad--a comprehensive real-world dataset for unsupervised anomaly detection,''
\newblock in {\em Proceedings of the IEEE/CVF conference on computer vision and pattern recognition}, 2019, pp. 9592--9600.

\bibitem{visa}
Yang Zou, Jongheon Jeong, Latha Pemula, Dongqing Zhang, and Onkar Dabeer,
\newblock ``Spot-the-difference self-supervised pre-training for anomaly detection and segmentation,''
\newblock in {\em European Conference on Computer Vision}. Springer, 2022, pp. 392--408.

\bibitem{mvtec3dad}
Paul Bergmann, Xin Jin, David Sattlegger, and Carsten Steger,
\newblock ``The mvtec 3d-ad dataset for unsupervised 3d anomaly detection and localization,''
\newblock {\em arXiv preprint arXiv:2112.09045}, 2021.

\bibitem{winclip}
Jongheon Jeong, Yang Zou, Taewan Kim, Dongqing Zhang, Avinash Ravichandran, and Onkar Dabeer,
\newblock ``Winclip: Zero-/few-shot anomaly classification and segmentation,''
\newblock in {\em Proceedings of the IEEE/CVF Conference on Computer Vision and Pattern Recognition}, 2023, pp. 19606--19616.

\bibitem{saa}
Matthew Baugh, James Batten, Johanna~P M{\"u}ller, and Bernhard Kainz,
\newblock ``Zero-shot anomaly detection with pre-trained segmentation models,''
\newblock {\em arXiv preprint arXiv:2306.09269}, 2023.

\bibitem{anovl}
Hanqiu Deng, Zhaoxiang Zhang, Jinan Bao, and Xingyu Li,
\newblock ``Anovl: Adapting vision-language models for unified zero-shot anomaly localization,''
\newblock {\em arXiv preprint arXiv:2308.15939}, 2023.

\bibitem{clipsurgery}
Yi~Li, Hualiang Wang, Yiqun Duan, and Xiaomeng Li,
\newblock ``Clip surgery for better explainability with enhancement in open-vocabulary tasks,''
\newblock {\em arXiv preprint arXiv:2304.05653}, 2023.

\end{thebibliography}

\end{document}